\documentclass[sigconf]{acmart}
\usepackage{soul}
\usepackage{url}
\usepackage{hyperref}
\usepackage[utf8]{inputenc}
\usepackage{caption}
\usepackage{graphicx}
\usepackage{amsmath}
\usepackage{amsthm}
\usepackage{booktabs}
\usepackage{algorithm}
\usepackage{algorithmic}
\usepackage{extarrows}
\usepackage{bm}
\usepackage{booktabs}
\usepackage{colortbl}
\usepackage{xcolor}         % colors
\usepackage{CJKutf8}
\usepackage{makecell}
\usepackage{amsmath,amsfonts,bm}

\def\eqref#1{equation~\ref{#1}}
\def\1{\bm{1}}

\DeclareMathAlphabet{\mathsfit}{\encodingdefault}{\sfdefault}{m}{sl}
\SetMathAlphabet{\mathsfit}{bold}{\encodingdefault}{\sfdefault}{bx}{n}

\newcommand{\Ls}{\mathcal{L}}

\newlength\savewidth\newcommand\shline{\noalign{\global\savewidth\arrayrulewidth\global\arrayrulewidth 1pt}\hline\noalign{\global\arrayrulewidth\savewidth}}

\definecolor{bluel}{RGB}{240,250,220}
\newcolumntype{a}{>{\columncolor{bluel}}c}
\usepackage{multirow}

\usepackage{pifont}
\newcommand{\cmark}{\ding{51}}%
\newcommand{\xmark}{\ding{55}}

\AtBeginDocument{%
  }

\copyrightyear{2023} 
\acmYear{2023} 
\setcopyright{acmlicensed}
\acmConference[MM '23]{Proceedings of the 31st
ACM International Conference on Multimedia}{October 29-November 3,
2023}{Ottawa, ON, Canada}
\acmBooktitle{Proceedings of the 31st ACM International Conference on
Multimedia (MM '23), October 29-November 3, 2023, Ottawa, ON, Canada}
\acmPrice{15.00}
\acmDOI{10.1145/3581783.3611788}
\acmISBN{979-8-4007-0108-5/23/10}

\acmSubmissionID{410}

\begin{document}

%%
%% The "title" command has an optional parameter,
%% allowing the author to define a "short title" to be used in page headers.
\title{Avatar Knowledge Distillation: Self-ensemble Teacher Paradigm with Uncertainty}

%%
%% The "author" command and its associated commands are used to define
%% the authors and their affiliations.
%% Of note is the shared affiliation of the first two authors, and the
%% "authornote" and "authornotemark" commands
%% used to denote shared contribution to the research.
\author{Yuan Zhang}
\authornote{This work was done during internship of Yuan Zhang in Alibaba DAMO Academy.}
\email{zhangyuan@stu.pku.edu.cn}
\affiliation{%
  \institution{Peking University}
  \city{Beijing}
  \country{China}
}

\author{Weihua Chen}
\email{kugang.cwh@alibaba-inc.com}
\affiliation{%
  \institution{Alibaba Group}
  \city{Beijing}
  \country{China}
}

\author{Yichen Lu}
\email{yicheng.lyc@alibaba-inc.com}
\affiliation{%
  \institution{Alibaba Group}
  \city{Beijing}
  \country{China}
}

\author{Tao Huang}
\email{huntocn@gmail.com}
\affiliation{%
  \institution{The University of Sydney}
  \city{Sydney}
  \country{Australia}
}

\author{Xiuyu Sun}
\authornote{Correspondence to: Xiuyu Sun.}
\email{xiuyu.sxy@alibaba-inc.com}
\affiliation{%
  \institution{Alibaba Group}
  \city{Beijing}
  \country{China}
}

\author{Jian Cao}
\email{caojian@ss.pku.edu.cn}
\affiliation{%
  \institution{Peking University}
  \city{Beijing}
  \country{China}
}

%%
%% By default, the full list of authors will be used in the page
%% headers. Often, this list is too long, and will overlap
%% other information printed in the page headers. This command allows
%% the author to define a more concise list
%% of authors' names for this purpose.
\renewcommand{\shortauthors}{Yuan Zhang et al.}

\begin{abstract}
    Knowledge distillation is an effective paradigm for boosting the performance of pocket-size model, especially when multiple teacher models are available, the student would break the upper limit again. However, it is not economical to train diverse teacher models for the disposable distillation. In this paper, we introduce a new concept dubbed \textbf{\textit{Avatars}} for distillation, which are the inference ensemble models derived from the teacher. Concretely, (1) For each iteration of distillation training, various Avatars are generated by a perturbation transformation. We validate that Avatars own higher upper limit of working capacity and teaching ability, aiding the student model in learning diverse and receptive knowledge perspectives from the teacher model. (2) During the distillation, we propose an \textbf{\textit{uncertainty-aware}} factor from the variance of statistical differences between the vanilla teacher and Avatars, to adjust Avatars' contribution on knowledge transfer adaptively. Avatar Knowledge Distillation (\textbf{\textit{AKD}}) is fundamentally different from existing methods and refines with the innovative view of unequal training. Comprehensive experiments demonstrate the effectiveness of our Avatars mechanism, which polishes up the state-of-the-art distillation methods for dense prediction without more extra computational cost. The AKD brings at most $\textbf{0.7 AP}$ gains on COCO 2017 for Object Detection and $\textbf{1.83 mIoU}$ gains on Cityscapes for Semantic Segmentation, respectively. Code is available at \url{https://github.com/Gumpest/AvatarKD}.
\end{abstract}

%%
%% The code below is generated by the tool at http://dl.acm.org/ccs.cfm.
%% Please copy and paste the code instead of the example below.
%%
\begin{CCSXML}
<ccs2012>
   <concept>
       <concept_id>10010147.10010178</concept_id>
       <concept_desc>Computing methodologies~Artificial intelligence</concept_desc>
       <concept_significance>500</concept_significance>
       </concept>
   <concept>
       <concept_id>10010147.10010257</concept_id>
       <concept_desc>Computing methodologies~Machine learning</concept_desc>
       <concept_significance>300</concept_significance>
       </concept>
 </ccs2012>
\end{CCSXML}

\ccsdesc[500]{Computing methodologies~Artificial intelligence}
\ccsdesc[300]{Computing methodologies~Machine learning}

%%
%% Keywords. The author(s) should pick words that accurately describe
%% the work being presented. Separate the keywords with commas.
\keywords{Model Compression, Knowledge Distillation, Avatars Mechanism}
%% A "teaser" image appears between the author and affiliation
%% information and the body of the document, and typically spans the
%% page.
% \begin{teaserfigure}
%   \includegraphics[width=\textwidth]{sampleteaser}
%   \caption{Seattle Mariners at Spring Training, 2010.}
%   \Description{Enjoying the baseball game from the third-base
%   seats. Ichiro Suzuki preparing to bat.}
%   \label{fig:teaser}
% \end{teaserfigure}

% \received{24 April 2023}
% \received[revised]{12 March 2009}
% \received[accepted]{5 June 2009}

% \settopmatter{printacmref=false} %remove ACM reference format

%%
%% This command processes the author and affiliation and title
%% information and builds the first part of the formatted document.
\maketitle

\section{Introduction}
\label{sec:intro}
% motivation
\begin{figure}[t]
    \centering
    \includegraphics[width=0.35\textwidth]{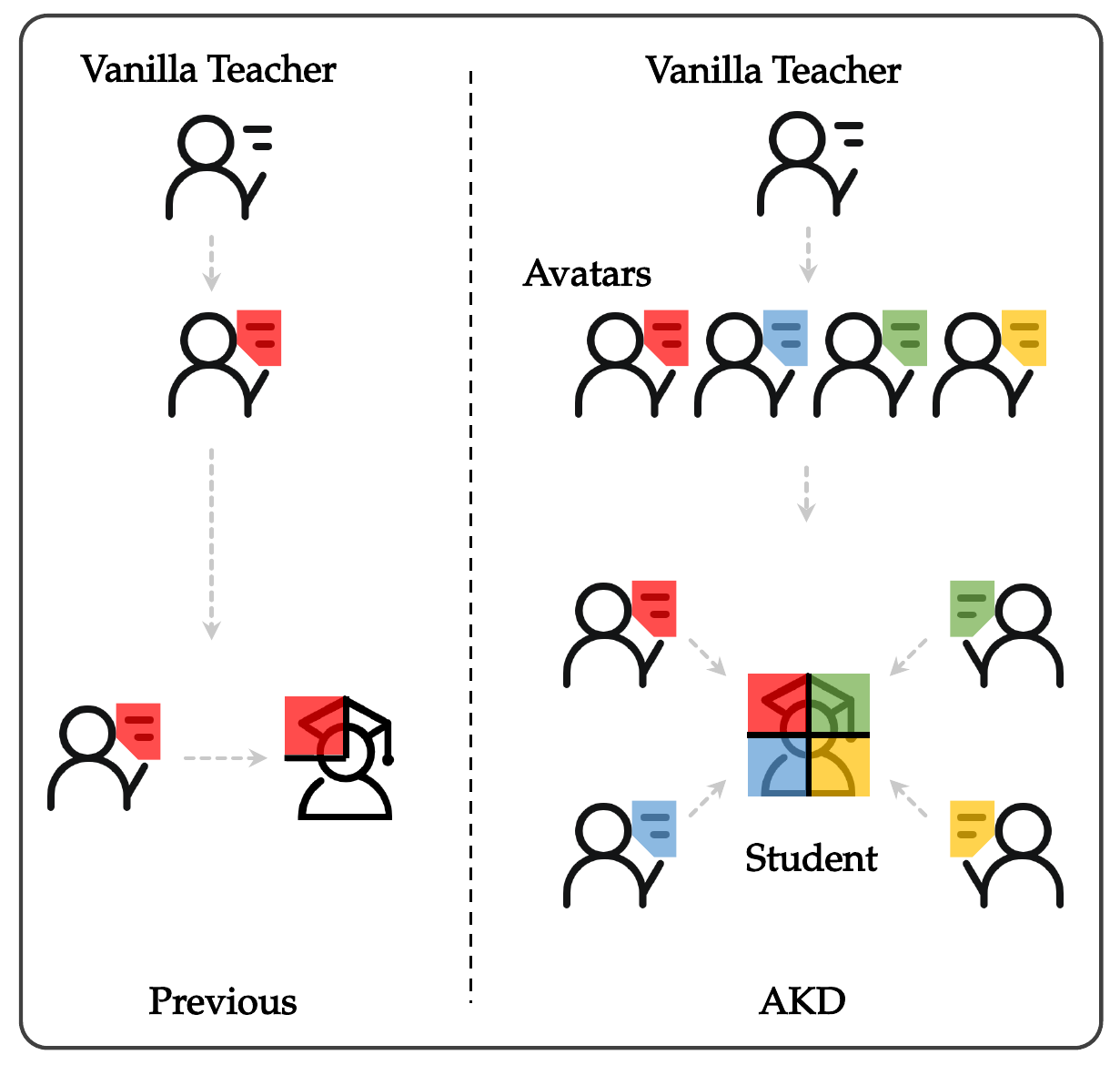}
    % \vspace{-1mm}
    \caption{\textbf{An illustration of Avatars distillation mechanism.} Avatars ensemble not only owns stronger performance, but also more exquisite teaching skills than the vanilla teacher. Best view in color.}
    \label{fig:motivation}
    % \vspace{-1mm}
\end{figure}

% distillation
Recent deep neural networks tend to grow deeper and wider for ultimate performance \cite{he2016deep, li2019selective, xie2017aggregated}. However, due to huge amounts of complicated parameters, such heavy models are clumsy and inefficient to deploy on edge devices, causing lower inference latency. To remedy the issue, knowledge distillation (KD) \cite{hinton2015distilling} has been proposed to inherit the dark knowledge from a heavy model (teacher) to a compact model (student) with similar structure, and make student generalize better than being trained alone.

% multiple teacher distillation
%Traditional knowledge distillation methods \cite{hinton2015distilling,romero2014fitnets} start from a single teacher. 
Taking the advance of model ensemble,
%As the wisdom of the masses exceeds that of the wisest individual, 
some recent methods \cite{liu2020adaptive,you2017learning, yuan2021reinforced} improve the distillation performance by
employing multiple teachers to teach the student simultaneously.
%with adaptive or fixed weights, for further improving its performance. 
The basic motivation behind these methods is that, compared to single teacher, multiple teachers can provide more various and informative supervisions to the student by generating diverse ``views'' of each given sample, and thus lead the student to generalize better~\cite{fukuda2017efficient,yuan2021reinforced}. However, obtaining such diverse teachers requires to train multiple teachers, which is computationally-expensive, uneconomic, and ungreen.

In this paper, we aim to solve the above problem by proposing Avatar Knowledge Distillation (AKD), which only utilizes one teacher to generate such diverse ``views'' of samples in multiple teachers. Concretely, with the deterministic network parameters of a teacher, AKD leverages an efficient Bayesian perturbation in Bayesian Neural Networks (BNNs)~\cite{blundell2015weight,kingma2015variational} to disturb the teacher features into multiple diverse features, which we call Avatars, to supervise the student. As alternates of multiple teachers, these Avatars can also provide multiple discriminate perspectives of data samples to enhance the student, as illustrated in Fig. \ref{fig:motivation}.

Nevertheless, it is inevitable for Avatars to bring extra noises into distillation due to the perturbation, while these noises could mislead the student from learning beneficial knowledge. Therefore, how to control the stability of training process and reduce the impact of noise is a vital question deserved to explore.
%The noise is called epistemic uncertainty in \cite{kendall2017uncertainties}.
%For the teacher who is sensitive to the noise, its generation is easily to be crashed, and the generated Avatars would lead to a poor result that may harm the distillation. Hence, how to control the influence of the generation and its Avatars on the distillation is an important iusse.
%When meeting multifarious samples, who deserves more attention to teach the student is a concern. 
%Early methods \cite{you2017learning,fukuda2017efficient} combine predictions from multiple teachers with the fixed weight assignment. 
Recently, some methods propose adaptive weight schema to balance the influence of multiple teachers in distillation. For example, AMTML-KD \cite{liu2020adaptive} calculates the weights based on latent factors, while CA-MKD \cite{zhang2022confidence} affirms that the teacher's predictions closer to the labels should be assigned to larger weights. However, such heuristic rules may not be ideal proxies to measure the contributions of each teacher, and therefore limit the distillation performance of multiple teachers.
In this paper, considering that our Avatars are generated through Bayesian perturbation, we can naturally model the noise as an uncertainty from the generation,
%we model the noise as an uncertainty from the generation, 
and 
leverage an uncertainty-aware factor to adjust the impact of Avatars on distillation. 
With the help of our uncertainty-aware weight, we could avoid the noise in Avatars damaging the knowledge transfer process.

% conclude
As analyzed above, we propose an \textbf{A}vatar \textbf{K}nowledge \textbf{D}istillation (AKD), involving multiple Avatars to ensemble distillation adaptively, as shown in Fig. \ref{fig:motivation}. Various Avatars are only generated from intermediate features, so that AvatarKD can be directly extended to existed methods on several vision tasks, including classification, object detection and semantic segmentation.
Without bells and whistles, our method can further improve baseline algorithms and achieve state-of-the-art performances consistently on detection and semantic segmentation benchmarks. In a nutshell, the contributions of this paper are:
\begin{enumerate} 
\item We design Avatar Knowledge Distillation (AvatarKD), a self-ensemble teacher paradigm, to enrich the diversity for distillation, without the cost of training multiple teachers.
The mechanism is simple but efficient, and orthogonal with other distillation methods.
\item An uncertainty-aware weight is modelled from the generation to measure the influence and stability of Avatars on distillation, which aids to weaken the negative effects from the noise in Avatars.
%We model Avatar teaching quality by introducing aleatoric uncertainty. %Receiving the student performance feedback, the Avatars with versatile appearance show different degree of reliability.
\item We verify the effectiveness of our method on object detection and semantic segmentation tasks via extensive experiments based on existed methods, achieving state-of-the-art performance.
%Our teacher avatar mechanism is simple but efficient. We verify the effectiveness of our method on detection and semantic segmentation tasks via extensive experiments based on existed methods, achieving state-of-the-art performance.
\end{enumerate} 

%------------------------------------------------------------------------

\section{Related Work}
\label{sec:revisiting}

\subsection{Knowledge Distillation}
As the seminal work by Hinton \cite{hinton2015distilling}, knowledge distillation is introduced to boost performance of tiny models. A line of works \cite{huang2022knowledge, ahn2019variational, jin2023video, huang2023knowledge, zhang2021distilling} has verified its potential. To extend its application to object detection, MIMIC \cite{li2017mimicking} proposed to prevent divergence between RPN output of teacher ans student via $ L_2 $ loss.
To filter massive less informative supervision in background areas, Wang \cite{wang2019distilling} transferred effective knowledge in sparse imitation regions.
To handle imbalanced supervision from foreground and background pixels, Zhang \cite{zhang2020improve} and Yang \cite{yang2022focal} encouraged student to learn from crucial pixels of foreground objects and the relationship between them. 
To separate informative feature in and out of bounding boxed, FRS~\cite{Du2021DistillingOD} presented feature-richness score to focus on important features.
To further investigate sophisticated dependence, MasKD \cite{huang2022masked} proposed receptive tokens to build pixel dependency masks for more effective distilling. Instead of spatial-wise, CWD~\cite{shu2021channel} exploited more semantical information in each channel with category-specific masks. For the sake of fusing different kinds of information, GID \cite{dai2021general} merged branches of feature/relation/response-based knowledge and developed an all-in-one framework. In most recent work MGD \cite{yang2022masked}, distillation was combined with self-supervision task, it implicitly guided student via recovering teacher output with masked student features, introducing another generative way in distillation.

\subsection{Multiple Teachers Distillation}
Based on knowledge distillation with single teacher, multiple teachers could provide supervision from other more perspectives of data and lead to more impressive performance. Prior works conducted such teaching with an ensembled teacher \cite{chebotar2016distilling} or with fixed weight average \cite{you2017learning,wu2019multi} from multiple teachers. To better handle complex scenarios, with regard to each individual instance, each teacher responses differently and should be assigned with different weight adaptively. Following such idea, Liu \cite{liu2020adaptive} proposed to adaptively learning multi-level knowledge from teachers. Furthermore, CA-MKD~\cite{zhang2022confidence} bases weight assignment strategy on confidence of teachers. 
AKDE~\cite{kwon2020adaptive} proposed that student should trust teachers that predict with lower entropy.
In AE-KD~\cite{du2020agree}, multiple teacher distillation is treated as a multi-objective optimization problem to encourage better optimization direction. Methods above are built on assumptions that better teacher supervision is related to higher confidence, lower entropy and similar gradients, etc. For better generalization, RL-KD~\cite{yuan2021reinforced} proposed to dynamically assign weights to multiple teachers with respect to different data with reinforcement learning.

However, as they named it, all of methods above are driven by multiple teachers, which is not economical and not practical in computational-limited scenarios. Expanding one teacher to its multiple avatars could provide a promising solution from another perspective. Because one teacher means more practical and multiple avatars bring more working capacity and teaching ability (as detailed in Sec.~\ref{sec:comparsion}), which seems to be a free lunch in distillation.

\subsection{Uncertainty Estimation in Computer Vision}
Uncertainty \cite{maddox2019simple} describes the robustness in model inference. Kendall \cite{kendall2017uncertainties} formulated two types of uncertainty, epistemic uncertainty and aleatoric uncertainty from data and model respectively. Estimating such uncertainty denoises the training procedure and encourages model to learn from reliable operation with less risks. And with this pronounced advantage, many works in face recognition \cite{shi2019probabilistic,zhang2021relative}, person ReID \cite{zheng2021exploiting}, semantic segmentation \cite{kendall2017uncertainties,zheng2021rectifying}, object detection \cite{wang2021data} and knowledge distillation \cite{zhang2020prime}, have justified uncertainty in computer vision.
Uncertainty-based methods focus on the noise-resistant environments. In a standard pipeline, they suppose that uncertainty in data or prediction is strongly related to quality of supervision. And uncertainty could stem from noise in mislabelling, ill pseudo-labels, inconsistent output and so on. After quantifying the noise and turning it into uncertainty, methods scale the objective function and put more attention on reliable data with lower uncertainty.

Different from existing work, to our best knowledge, our work is the first exploration about uncertainty-aware teachers (Avatars) from the generation in distillation. In this paper, we quantify the uncertainty of each element in feature maps and dynamically assign weights to distillation term in the most refined way.

%------------------------------------------------------------------------

%------------------------------------------------------------------------

\section{Proposed Approach}
\begin{figure}
    \centering
    \includegraphics[width=0.38\textwidth]{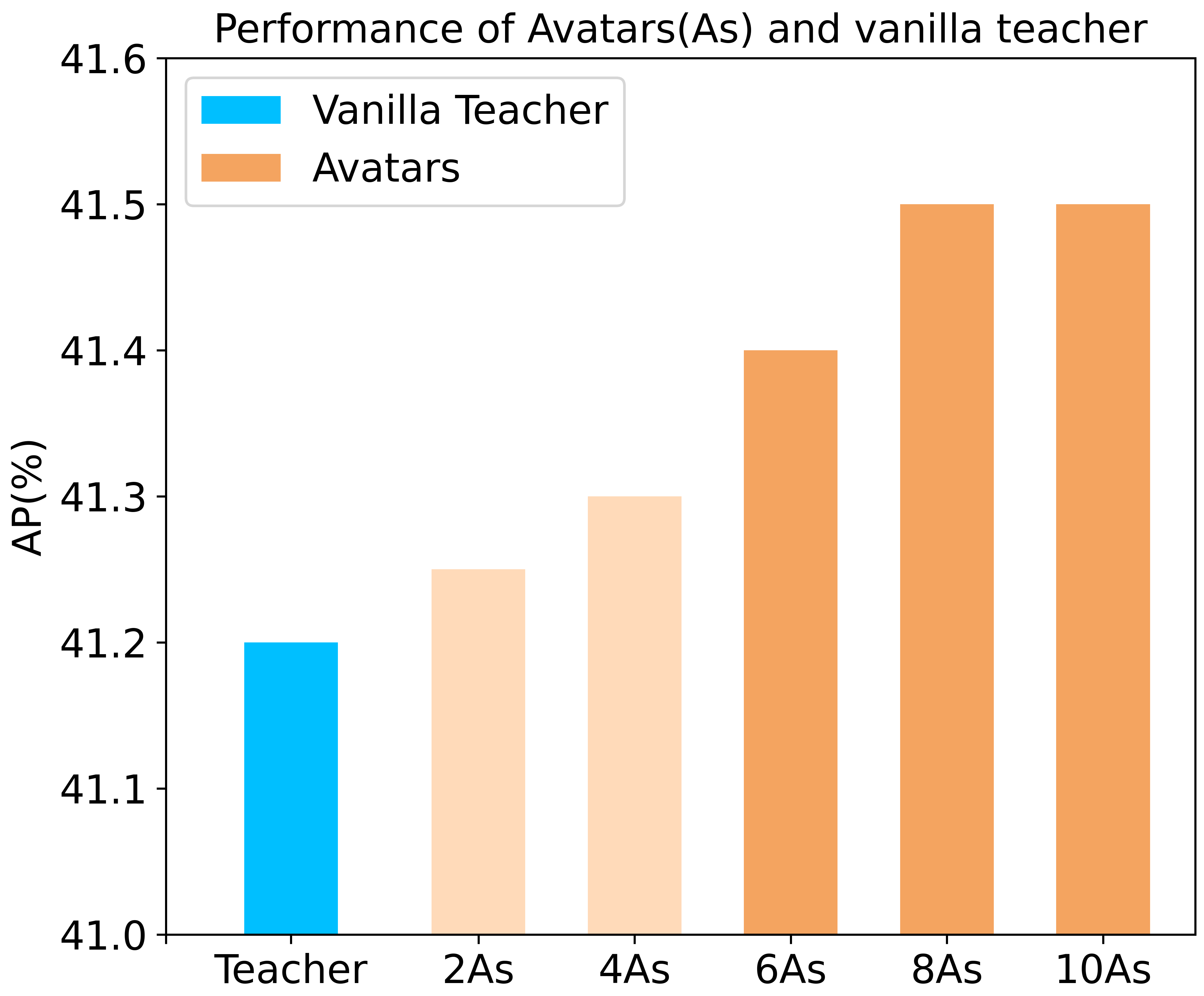}
    %\vspace{-2mm}
    \caption{\textbf{Experiment on the performance of ensemble Avatars based on \textit{RetinaNet-X101} detector on COCO 2017.}}
    \label{fig:3.1}
    \vspace{-1mm}
\end{figure}

\begin{figure*}[t]
    \centering
    \includegraphics[width=0.99\textwidth]{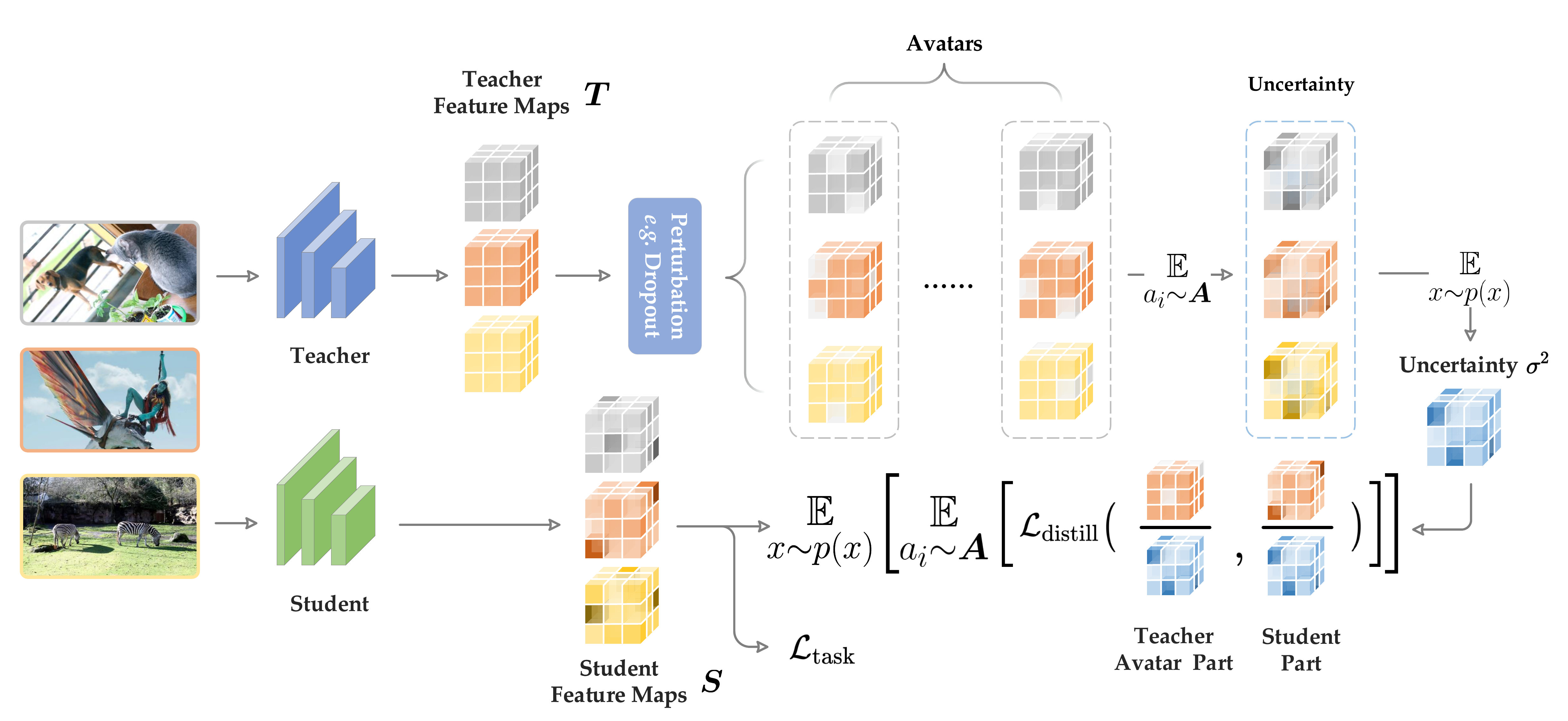}
    %\vspace{-4mm}
    \caption{\textbf{Overview of our AKD framework.} Teacher feature maps separated into $ \bm{N} $ avatars, containing diverse perspective knowledge. To measure the quality of each randomly perturbed avatars, uncertainty is introduced and calculated among all avatars of training data. Combining the merged uncertainty to scale affects adaptively on dimensions (\textit{e.g.} channels, spatial grids), distillation process is conducted between student and teacher avatars.}
    \label{fig:main}
    %\vspace{-1mm}
\end{figure*}

\subsection{Multiple Avatars are Better than a Teacher}
\label{sec:comparsion}
% generation
% The Avatar $\bm{A}_{i}(x)$ is generated by the perturbation $\boldsymbol{P}$, stemming from the teacher model $\boldsymbol{T}(x)$. The generation can be formulated as:
The Avatar $\boldsymbol{A_i}$ is generated by the perturbation $\boldsymbol{P}$, whose feature maps $ \bm{F}^{(a_{i})} $ stem from feature maps $ \bm{F}^{(t)} $ of teacher model $\boldsymbol{T}$. The generation can be formulated as:
\begin{equation}
  \bm{F}^{(a_{i})} = \bm{P}(\bm{F}^{(t)}),
\end{equation}
where $\bm{F}^{(a_i)}\in\mathbb{R}^{C\times H\times W}$ and $\bm{F}^{(t)}\in\mathbb{R}^{C\times H\times W}$ denote the feature maps the student would be imitated (\textit{eg}, the outputs of feature pyramid network~\citep{lin2017feature} (FPN) in object detection tasks). Inspired by~\cite{kendall2017uncertainties}, we use Bayesian Neural Network (BNN)~\cite{blundell2015weight,kingma2015variational} as the perturbation $\bm{P}$ to produce our Avatars. 

In our paper, we employ the single dropout layer with ratio $m_i$ = 0.1 instead of Multi-Layer Perception (MLP) to form the BNN, which is simple but effective, to avoid extra computational cost. Notably, other augmentation methods can also replace the dropout to function as the perturbation $\bm{P}$, providing disturbance.

During Avatars distillation, one typical manner is to mimic the tensor pixel-wisely \citep{romero2014fitnets, chen2017learning}. The vanilla Avatars distillation can be compeleted via Avatars ensemble model $\bm{A}$, when each avatar makes the equal contribution to the supervision: 
\begin{equation} \label{eq:mimic}
    \Ls_\mathrm{distill}(\bm{A}, \bm{S}) = \frac{1}{k} \sum_i^{k} \frac{1}{HWC} \left\|\bm{F}^{(a_i)} - \phi(\bm{F}^{(s)})\right\|_2^2 ,
\end{equation}
where $\bm{F}^{(s)}$ is the feature maps of the student model, $k$ is the amount of Avatars and $\phi$ is a linear projection layer to adapt $\bm{F}^{(s)}$ to the same resolution as $\bm{F}^{(a_i)}$.

Next, we will demonstrate the reason that why multiple Avatars are better than a teacher. Working capacity means the accuracy of the teacher model on corresponding tasks, while teaching ability means the accuracy of the student model supervised by the teacher model. In Knowledge Distillation, better working capacity cannot be completely equivalent to better teaching ability, because when the gap between the teacher and student is too large, it will cause a negative effect on distillation. Therefore, we prove the advantage of our Avatars in both aspects.

% s1 s2 s3 --> (wbf) > t1 (personal ability)
\textbf{Working capacity.} We first validate the Avatars ensemble model owns a better working capacity on specific tasks. As shown in Fig. \ref{fig:3.1}, we conduct the experiment on the performance of ensemble\footnote{Weighted Box Fusion (WBF) \cite{solovyev2021weighted} is used to ensemble Avatars' outputs} Avatars and teacher on COCO dataset with RetinaNet-X101. When the number of Avatars exceeds one, their ensemble performance is constantly superior to teacher's. Ensemble model always owns richer feature representation and stronger robustness compared with the single one, which is worthy of imitation.

% + dropout --> (distill) > t1 (teaching ability)
\textbf{Teaching ability.} Besides, we further prove that the Avatars ensemble model affords more exhaustive guide for the student. Various data distribution in Avatars makes the student learn the diverse perspective of feature maps, and the unequal distillation ensures the student receive high-quality knowledge. As the experiments conducted in Tab. \ref{tab:moti},we train the student \textit{RetinaNet-R50} with Avatars (equal weights ensemble) and vanilla teacher \textit{RetinaNet-X101} using 2$\times$ schedule, and adopts CWD (based on KL) and MGD (based on MSE) distillation on feature. Avatars constantly bring extra gain (0.2 AP) on distillation, proving that choosing multiple Avatars is a better schema.

\begin{table}[t]
    \centering
    \begin{minipage}[c]{0.50\textwidth}
	\renewcommand\arraystretch{1.4}
	\setlength\tabcolsep{2.8mm}
	\centering
	%\vspace{-4mm}
	\caption{\textbf{Performance of the distilled student under the supervision of Avatars (equal weights ensemble) and vanilla teacher, respectively.}}
     %\vspace{-2mm}
	\label{tab:moti}
	%\footnotesize
	\begin{tabular}{c|c|c}
	    \shline
	    Method & Supervisor & AP(\%) \\
	    \shline
        CWD (KL) & Vanilla teacher & 40.8 \\
	    CWD (KL) & Avatars & \textbf{41.0} \\
	    \shline
        MGD (MSE) & Vanilla teacher & 41.0 \\
	    MGD (MSE) & Avatars & \textbf{41.2} \\
	    \shline
	\end{tabular}
     %\vspace{-2mm}
	\end{minipage}\hfill
\end{table}

The generation of Avatars is presented in Fig.~\ref{fig:main}. The perturbation applies on the feature maps from the teacher, and produce $k$ Avatars in each iteration, which boost the performance of student.

%As shown in Tab \ref{tab:multi-tea}, we conduct the experiments on the performance of ensemble (WBF \cite{solovyev2021weighted}) Avatars on COCO with RetinaNet-X101, to compare multiple Avatars with the teacher. When we generate two Avatars, their ensemble AP is just the same as teacher's. Nevertheless, when the number of Avatars is more than four, their ensemble performance is always superior to the teacher's. As the proverb says, two heads are better than one.

%The reason for why multi-Avatars are better than the single teacher is that the ensembled model owns richer feature representation and more robustness, while the generation cost is low.

\subsection{The Uncertainty-Aware Weight}
\label{sec:uncertainty}

However, when the Avatars are produced, the perturbation not only brings diversity into the Avatars, but also involves the noise that may drift the Avatar from the original teacher and even lead to a lower performance. The noise and the diversity is a tricky balance for the Avatars. 
%The generation process may even crashed when the original teacher is too sensitive to the noise. 
Hence, we have to find a way to watch the generation, and reduce Avatars' effects on distillation when the noise impact becomes severe.
%In real-world scenarios, the observation of one object is perturbed by noise $ \bm{n}$. 

During the generation, there exists disagreement between a specific Avatar output $ \bm{F}^{(a_{i})} $ and the teacher output $ \bm{F}^{(t)} $, \textit{i.e.}, $ \bm{F}^{(a_{i})} = \bm{F}^{(t)} + \bm{n} $, caused by the perturbation. Assuming that the noise $ \bm{n} $ follows a normal distribution $ \mathcal{N}(0, \bm{\sigma}_i) $, the posterior of each sample can be formulated as:
\begin{equation}
    p( \bm{F}^{(a_{i})} | x ) =  \frac{1}{\sqrt{2 \pi \bm{\sigma}_{ix^{2}}}} \exp \left( -\frac{\big( \bm{F}^{(a_{i})} - \bm{F}^{(t)} \big)^2 }{ 2 \bm{\sigma}_{ix}^{2}} \right)
\end{equation}
where $ \bm{\sigma}_{ix} $ is the variance for input data $x$ on Avatar $i$. In~\cite{kendall2017uncertainties}, $ \bm{\sigma}_{ix} $ is treated as epistemic uncertainty, which is employed for measuring the potential variance of the model. The Avatar output $ \bm{F}^{(a_{i})} $, the teacher output $ \bm{F}^{(t)} $ and the variance $ \bm{\sigma}_{ix} $ are all given in form of tensor, \textit{i.e.}, $ \{ \bm{F}^{(a_{i})}, \bm{F}^{(t)}, \bm{\sigma}_{ix} \} \in \mathbb{R}^{{C} \times {H} \times {W}} $. After traversing over all possible samples, posterior turns into the equation below:
\begin{equation}
    p( \bm{F}^{(a_{i})} | x ) = \mathbb{E}_{x \sim p(x)} \left[ \frac{1}{\sqrt{2 \pi \bm{\sigma}_{ix^{2}}}} \exp \left( -\frac{\big( \bm{F}^{(a_{i})} - \bm{F}^{(t)} \big)^2 }{ 2 \bm{\sigma}_{ix}^{2}} \right) \right], 
    \label{equ:posteriori}
\end{equation}
$p( \bm{F}^{(a_{i})} | x )$ is the posterior for Avatar $i$. To minimize the negative log likelihood of posterior, the objective is achieved across all Avatars:
\begin{equation}
    \begin{aligned}
        \mathcal{L}&_{ \text{uncertainty}}(x) = - \log p( \bm{A} | x ) \\ 
        & = \mathbb{E}_{ x \sim p(x) } \left[ \mathbb{E}_{ a_i \sim \bm{A} } \left[ \frac{1}{2} \log 2 \pi  \bm{\sigma}_{ix}^2 + \frac{\big( \bm{F}^{(a_{i})} - \bm{F}^{(t)} \big) ^ 2 }{ 2 \bm{\sigma}_{ix}^2 } \right] \right] \\
        & \propto \mathbb{E}_{ x \sim p(x) } \left[ \mathbb{E}_{ a_i \sim \bm{A} } \left[ \underbrace{  \log \bm{\sigma}_{ix}^2 + \frac{\big( \bm{F}^{(a_{i})} - \bm{F}^{(t)} \big) ^ 2 }{ \bm{\sigma}_{ix}^2 } }_{ \text{local} } \right] \right]. \\
    \end{aligned}
\end{equation}
For the local uncertainty term, it will reach analytical minimum, when we let $\bm{\sigma}_{ix}^2 = \big( \bm{F}^{(a_{i})} - \bm{F}^{(t)} \big)^2 $. 

Considering the whole generation, the optimal uncertainty $ \bm{\sigma} $ will converge to
\begin{math}
 \bm{\sigma}^2 = \mathbb{E}_{x \sim p(x)} \left[ \mathbb{E}_{ a_i \sim \bm{A} } \left[ ( \bm{F}^{(a_{i})} - \bm{F}^{(t)} )^2 \right] \right]
\end{math}
 across all Avatars and all data. 
Here with the dropout operation as an example, \textit{i.e.}, $ \bm{F^{(a_i)}} = \text{dropout}(\bm{F^{(t)}}, 1-m_i) $, where $ m_i $ is the corresponding dropout ratio, the uncertainty $ \bm{\sigma} $ of the whole generation could be further simplified:
\begin{equation}
    \begin{aligned}
        { \bm{\sigma} }^2
        %& = \mathbb{E}_{x \sim p(x)} \left[ \bm{\sigma}_{x}^2 \right] \\
        & = \mathbb{E}_{x \sim p(x)} \left[ \mathbb{E}_{ a_i \sim \bm{A} } \left[ ( \bm{F}^{(a_{i})} - \bm{F}^{(t)} )^2 \right] \right]  \\
        & = \mathbb{E}_{x \sim p(x)} \left[ {m_i^2} \left[ \bm{F}^{(t)} \right]^2 \right] \\
        & \quad \xlongequal{ \bm{\mu}_{x \sim p(x)} \left[ \bm{F}^{(t)} \right] = \bm{0} } m_i^2 \times \text{var}_{x \sim p(x)} \left[ \bm{F}^{(t)} \right] \\
        & \propto \text{var}_{x \sim p(x)} \left[ \bm{F}^{(t)} \right],
    \end{aligned}
    \label{eq:sigma}
\end{equation}
From Eq.~\ref{eq:sigma}, it can be found that the uncertainty (the noise brought by the perturbation) is directly proportional to the dropout ratio term $ m_i^2 $ and the statistical variance of the teacher across all data $\text{var}_{x \sim p(x)} \left[ \bm{F}^{(t)} \right]$, under the condition of $ \bm{\mu}_{x \sim p(x)} \left[ \bm{F}^{(t)} \right]=\bm{0} $. 
It is noteworthy that the condition $ \bm{\mu}_{x \sim p(x)} \left[ \bm{F}^{(t)} \right]=\bm{0} $ could be easily achieved by estimating and subtracting mean of global features. In this paper, we involve a Batch Normalization (BN) Layer after the teacher feature maps $T$ before the perturbation $P$ (\textit{i.e.,} dropout) to achieve $ \bm{\mu}_{x \sim p(x)} \left[ \bm{F}^{(t)} \right]=\bm{0} $.
And the dropout ratio term $ m_i^2 $ can be set to a fixed value and controlled manually. So that the uncertainty is proportional to the statistical variance of the teacher across all data $\text{var}_{x \sim p(x)} \left[ \bm{F^{(t)}} \right]$. 
% \textcolor{red}{[TODO]}Based on this observation, we take the moving average variance from the BN layer as $\text{var}_{x \sim p(x)} \left[ \bm{T(x)} \right]$.
Finally, the uncertainty is obtained by Eq.~\ref{eq:sigma} with the form of variance of each position in tensor $ \mathbb{R}^{{C} \times {H} \times {W}} $ and could be further merged (\textit{e.g.} in channel-wise or spatial-wise) to get a more stable form. Detailed discussion is reported in Sec.~\ref{sec:merge}.

To plug the uncertainty-aware weight, the eventual distillation objective $\mathcal{L}_{\text{AKD}}$ is formulated as:
\begin{equation}
    \begin{aligned}
    \mathcal{L}_{\text{AKD}}(\bm{S}, \bm{A}) 
    %& = \mathbb{E}_{\mathbb{C}, \mathbb{H}, \mathbb{W}} \left[ \mathcal{L}_{\text{distill}}( \lambda\bm{T}_{c, h, w}, \lambda\bm{A}_{c, h, w}) \right] \\
    % & = \mathbb{E}_{a_i} \left[ \mathbb{E}_{{C}, {H}, {W}} \left[ \mathcal{L}_{\text{distill}}( \frac{\phi(\bm{F}^{(s)})_{c, h, w}}{\bm{\sigma}_{c, h, w}}, \frac{\bm{F}^{(a_i)}_{c, h, w}}{\bm{\sigma}_{c, h, w}}) \right] \right]
    & = \frac{1}{k} \sum_i^{k} \frac{1}{HWC} \left\|\frac{\bm{F}^{(a_i)}_{c, h, w}}{\bm{\sigma}_{c, h, w}} - \frac{\phi(\bm{F}^{(s)})_{c, h, w}}{\bm{\sigma}_{c, h, w}}\right\|_2^2 
    \end{aligned}
\end{equation}

\subsection{Discussion}

\textbf{Uncertainty of Whole Feature Maps.}
In previous works on uncertainty~\cite{kendall2017uncertainties,zhang2020prime}, typically, a scalar is estimated to measure the noise in each sample and put more weight to cleaner data. In our approach, uncertainty is a tensor to reflect the variance of whole feature maps. Due to the statistics from global feature distribution, attention is put more precisely on different positions and induces better performance of student networks.
\begin{figure}[t]
    \centering
    \includegraphics[width=0.8\linewidth]{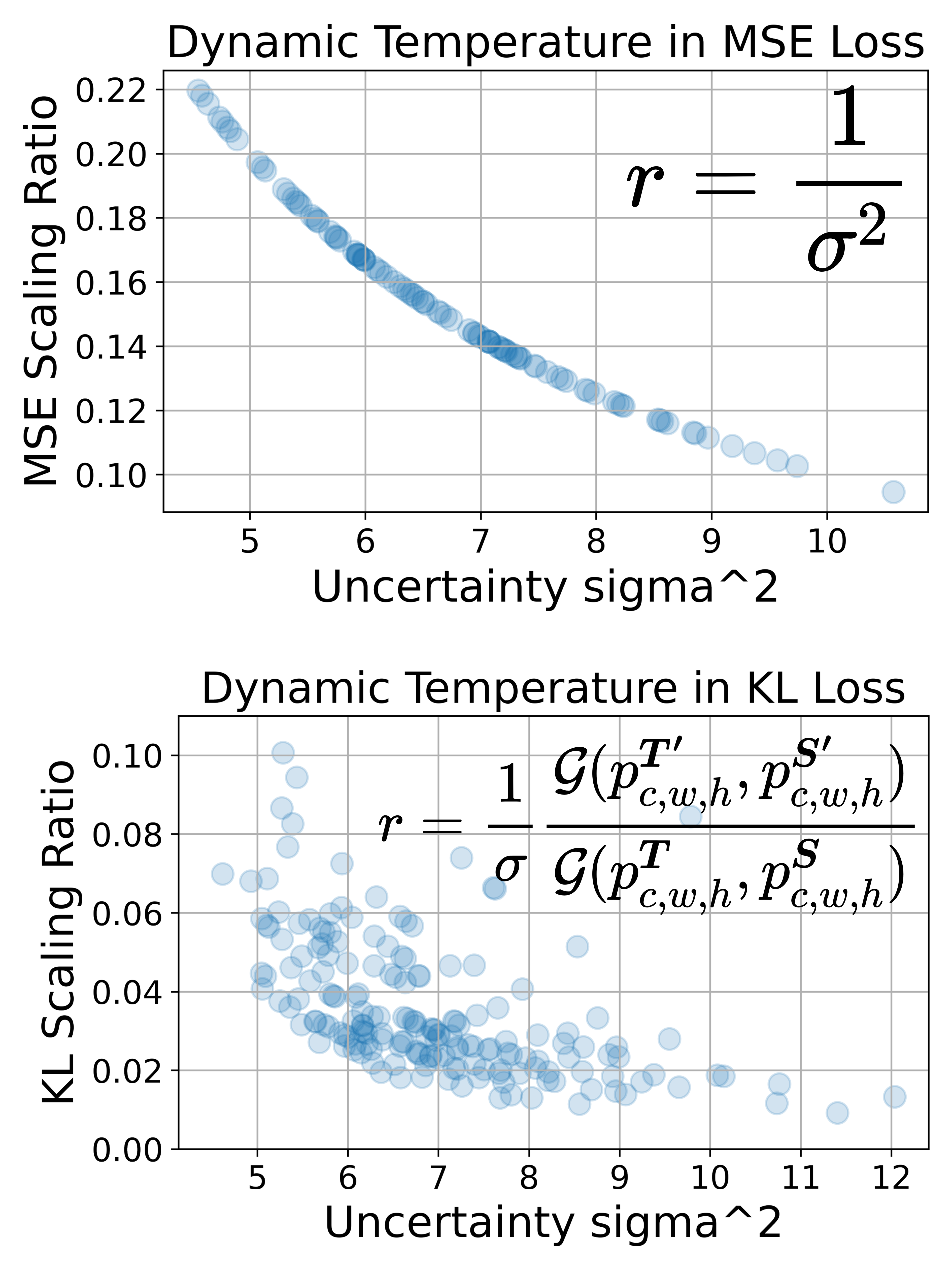}
    %\vspace{-2mm}
    \caption{Impact of uncertainty on gradients. We plot scatters: ratio of gradient \textit{v.s.} uncertainty, where ratio of gradient could be described as $ r = {\text{grad}_{\text{w/ } \sigma}} / {\text{grad}_{\text{w/o } \sigma} } $. We take MSE loss (Left) and KL loss (Right) as examples and their ratio formulas are pinned on top right corners. Note that $ \mathcal{G}(\cdot) $ in KL plot is formulated in Eq.~\ref{eq:kl_ratio}.}
    \label{fig:grad_scatter}
    \vspace{-2mm}
\end{figure}

\noindent
\textbf{Dynamic and General Temperature.} In vanilla distillation~\cite{hinton2015distilling}, temperature is proposed to produce a softer distribution over classes. In normal cases, distribution over classes could be described in single dimension, and a fixed temperature is enough to balance between details with lower probability and majority with higher probability. However, when it comes to distilling towards features in detection, more dimensions make it more troublesome to handle the variance in different dimensions. Over-softened features lose attention on foreground and under-softened features lead to ignorance of subtle changes. To relieve such flaw, our uncertainty could be viewed as dynamic temperature towards feature (or feature block) in different position. Besides, the concept of temperature is more general with uncertainty. It will not limited in form of KL divergence. In terms of gradients, the general form is written as:
\begin{equation}
    \frac{ \partial \mathcal{L} }{ \partial \bm{F}^{(s)}_{c, h, w} } = \frac{ \partial \mathcal{L}_{\text{distill}}( \bm{F}^{(t)}_{c, w, h} / \bm{\sigma}_{c, w, h}, \bm{F}^{(s)}_{c, w, h} / \bm{\sigma}_{c, w, h} )}{ \bm{\sigma}_{c, h, w} \; \partial \bm{F}^{(s)}_{c, h, w} }.
\end{equation}
Here, we take two types of widely-used loss function, \textit{i.e.}, MSE and KL, as examples:
\begin{equation}
    \frac{ \partial \text{ \tt{MSE}} ( \frac{ \bm{F}^{(t)}_{c, h, w} }{ \bm{\sigma}_{c, h, w} }, \frac{ \bm{F}^{(s)}_{c, h, w} }{ \bm{\sigma}_{c, h, w} } )}{ \partial \bm{F}^{(s)}_{c, h, w} } = \frac{ 2 (\bm{F}^{(s)}_{c, h, w} - \bm{F}^{(t)}_{c, h, w}) }{ \bm{\sigma}_{c, h, w}^2 } ,
\end{equation}
\begin{equation}
    \begin{aligned}
        \frac{ \partial \text{ \tt{KL}} ( p_{c, w, h}^{\bm{T}}, p_{c, w, h}^{\bm{S}} )}{ \partial \bm{S}_{c, h, w} } = \frac{\mathcal{G}(p_{c, w, h}^{\bm{T}}, p_{c, w, h}^{\bm{S}})}{\sigma_{c, h, w}}, \\
        \mathcal{G}(p_{c, w, h}^{\bm{T}}, p_{c, w, h}^{\bm{S}}) = p_{c, w, h}^{\bm{S}} \textstyle\sum_{i, j, k}{p_{i, j, k}^{\bm{T}}} - p_{c, h, w}^{\bm{T}} ,
    \end{aligned}
    \label{eq:kl_ratio}
\end{equation}
where $ p_{c, h, w}^{\bm{T}/\bm{S}} = \text{\tt{softmax}}(\bm{F}^{(t)/(s)}_{c, h, w} / \bm{\sigma}_{c, h, w} ) $. In formulas, uncertainty plays the role of temperature and gradients are all scaled adaptively as illustarted in Fig.~\ref{fig:grad_scatter}. 

% Completely the same is not necessarily correct, because limited capacity of student networks, its response to specific pattern cannot reach the as high level as teacher does. 

% uncertainty information mainly comes from . In other words, the sigma measures the uncertainty of each position in feature maps, not the uncertainty of feature map of the specific image.

%------------------------------------------------------------------------

\section{Experiments}
The AKD is a novel distillation mechanism that can easily be applied to different models for various tasks. In this section, to verify the compatibility of our approach on vision tasks (including classification, object detection and semantic segmentation), we plug AKD into three advanced feature-based distillation schema (including CWD \cite{shu2021channel} from ICCV 2021, MGD \cite{yang2022masked} from ECCV 2022 and MasKD \cite{huang2022masked} from ICLR 2023) based on MMRazor \cite{2021mmrazor}.

\subsection{Classification}
\subsubsection{\textbf{Datasets.}}
We validate our Avatars distillation efficacy on ImageNet-1K \cite{ImageNet} for classification, which contains 1000 object categories. We use the 1.2 million images for training and evaluate the student networks with accuracy (Acc) on 50k images.

\subsubsection{\textbf{Network architectures.}}
We follow MGD \cite{yang2022masked}, including homogeneous and heterogeneous distillation. Conduct comprehensive experiments on different detection frameworks, the one distillation setting is from ResNet-R34 \cite{he2016deep} to ResNet-R18, the other is from ResNet-R50 to MobileNetV2-R18 \cite{sandler2018mobilenetv2}.

\label{sec:cls-im}

\subsubsection{\textbf{Implementation Details.}} For the classification task, the feature maps of Avatars source from last feature map in the backbone of teacher. All the models are trained with the official strategies of 100 epochs schedule in MMClassification \cite{2023mmpretrain} with SGD optimizer based on Pytorch, where the momentum is 0.9 and the weight decay is 1e-4. We initialize the learning rate to 0.1 and decay it for every 30 epochs. This setting is based on 8 V100 GPUs.

\subsubsection{\textbf{Experimental results}}
\begin{figure}
    \centering
    \includegraphics[width=0.45\textwidth]{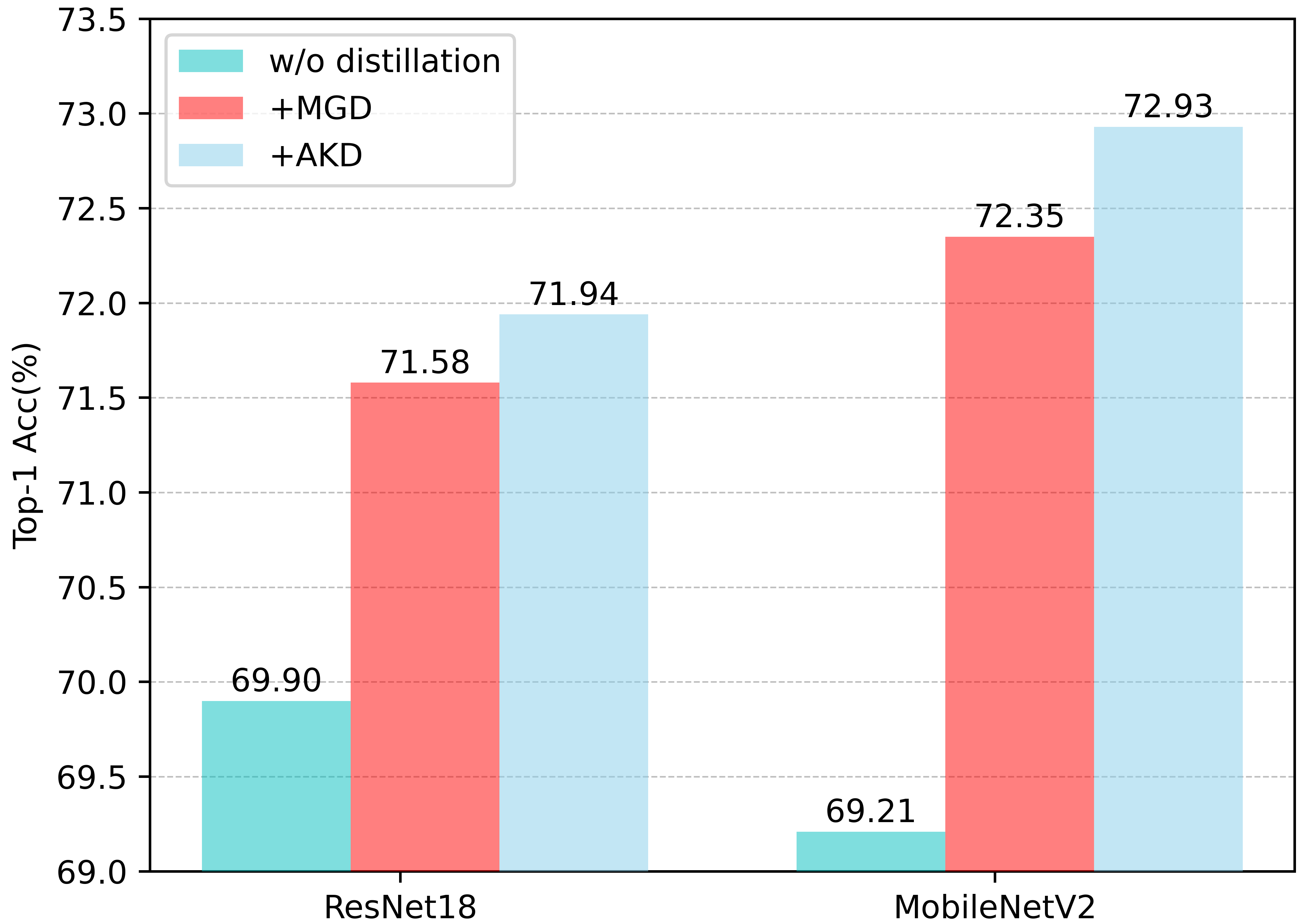}
    % \vspace{-10mm}
    \caption{\textbf{Performance of various KD strategies for classification (Top-1).}}
    \label{fig:cls}
    \vspace{-1mm}
\end{figure}

We transfer feature-based knowledge from backbone instead of logits-based knowledge. As shown in Fig. \ref{fig:cls}, compared with only by MGD distillation, the student ResNet-R18 and MobileNet-R18 gain 0.36 and 0.58 Top-1 accuracy improvement with our method, respectively. Various Avatars with adaptive weights further provide more robust and receptive feature maps information for the student.

\subsection{Object detection}
\subsubsection{\textbf{Datasets.}}
We validate our Avatars distillation efficacy on MS COCO detection dataset \cite{lin2014microsoft} for object detection, which contains 80 object classes. We evaluate the student networks with average precision (AP) on val2017 set.

\subsubsection{\textbf{Network architectures.}}
we conduct comprehensive experiments on different detection frameworks, including two-stage models \cite{ren2015faster}, anchor-based one-stage models \cite{lin2017focal}, and anchor-free one-stage models \cite{tian2019fcos,yang2019reppoints} to show our method effectiveness. 

\label{sec:od-im}

\subsubsection{\textbf{Implementation Details.}}
For the detection task, the feature maps of Avatars source from feature maps in the neck of teacher. We adopt ImageNet pre-trained backbones during training and inheriting strategy following previous KD works \cite{yang2022masked}. All the models are trained with the official strategies (SGD, weight decay of 1e-4) of 2X schedule in MMDetection \cite{chen2019mmdetection}. We run all the models on 8 V100 GPUs.

\subsubsection{\textbf{Experimental Results on Baseline Avatars}}
For thoughtful experiments, we first follow MasKD, and adopt multiple detection frameworks for our baseline settings, including two-stage detector Faster-RCNN, one-stage detector RetinaNet, and anchor-free detector FCOS. We take ResNet-101 (R101) backbone as the teacher network, with ResNet-50 (R50) as the student. 

As shown in Tab. \ref{tab:basedet}, our method indeed can further boost the performance of the two baseline algorithms. From the results that we report, Avatar mechanism steadily brings an increase of \textbf{0.3+ AP} gain (at most \textbf{0.5}) for CWD and MasKD, while under the same training setting. The result indicates that our method fits for different detection frameworks. Besides, the uncertainty-aware weight in AKD is suitable for both KL (average 0.40 AP gain on CWD) and MSE (average 0.33 AP gain on MasKD) loss function.

\definecolor{bluel}{RGB}{240,250,220}
\begin{table}[t]
    \centering
    % \hspace{2mm}
    \begin{minipage}[t]{0.50\textwidth}
    \renewcommand\arraystretch{1.35}
	\setlength\tabcolsep{0.35mm}
	\centering
	\caption{\textbf{Object detection performance with Avatars in baseline settings on COCO val set.} \textit{CM RCNN}: Cascade Mask RCNN.}
     %\vspace{-2mm}
	\label{tab:basedet}
	%\footnotesize
	\begin{tabular}{l|ccc|ccc}
	    \shline
	    Method & AP & AP$_{50}$ & AP$_{75}$ & AP$_{S}$ & AP$_{M}$ & AP$_{L}$\\
	    \shline
	    \scriptsize{T: Faster RCNN-R101} & 39.8 & 60.1 & 43.3 & 22.5 & 43.6 & 52.8 \\
	    \scriptsize{S: Faster RCNN-R50} & 38.4 & 59.0 & 42.0 & 21.5 & 42.1 & 50.3\\
        CWD~\cite{shu2021channel} & 40.5 & 60.8 & 44.0 & 23.3 & 44.6 & 53.4 \\
	    \rowcolor{bluel}
	    + AKD & \textbf{40.8} &  \textbf{61.0} & \textbf{44.7} & \textbf{23.4}& \textbf{44.6} & \textbf{53.9}\\
        MasKD~\cite{huang2022masked} & 40.6 & 60.7 & 44.2 & 23.0 & 44.0 & 53.9\\
	    \rowcolor{bluel}
	    + AKD & \textbf{40.9} & \textbf{60.9} & \textbf{44.7} & \textbf{23.3}& \textbf{44.5} & \textbf{54.1}\\
	    \shline

	    T: RetinaNet-R101 & 38.9 & 58.0 & 41.5 & 21.0 & 42.8 & 52.4\\
	    S: RetinaNet-R50 & 37.4 & 56.7 & 39.6 & 20.0 & 40.7 & 49.7 \\
        CWD~\cite{shu2021channel} & 39.8 & 59.0 & 42.6 & 21.7 & 43.5 & 53.4\\
	    \rowcolor{bluel}
	    + AKD & \textbf{40.2} & \textbf{59.1} & \textbf{42.8} & \textbf{22.2}& \textbf{44.0} & \textbf{54.7}\\
        MasKD~\cite{huang2022masked} & 39.9 & 59.0 & 42.7 & 21.4 & 43.6 & 53.8\\
	    \rowcolor{bluel}
	    + AKD & \textbf{40.3} & \textbf{59.4} & \textbf{43.0} & \textbf{21.8}& \textbf{44.1} & \textbf{54.2}\\
	    \shline

	    T: FCOS-R101 & 40.8 & 60.0 & 44.0 & 24.2 & 44.3 & 52.4\\
	    S: FCOS-R50 & 38.5 & 57.7 & 41.0 & 21.9 & 42.8 & 48.6\\
        CWD~\cite{shu2021channel} & {42.4} & {60.9} & {45.9} & {25.8} & {46.8} & {53.6} \\
	    \rowcolor{bluel}
	    + AKD & \textbf{42.9} & \textbf{61.3} & \textbf{46.5} & \textbf{26.6}& \textbf{46.8} & \textbf{54.8}\\
        MasKD~\cite{huang2022masked} & 42.2 & 60.9 & 45.6 & 26.2 & 46.4 & 53.6\\
	    \rowcolor{bluel}
	    + AKD & \textbf{42.5} & \textbf{61.2} & \textbf{46.0} & \textbf{26.5}& \textbf{46.8} & \textbf{54.0}\\
	    \shline
     
	\end{tabular}
    \vspace{-2mm}
    \end{minipage}
\end{table}

\subsubsection{\textbf{Experimental Results on Stronger Avatars}}
Following CWD and MGD, we conduct experiments on stronger teacher detectors, including two-stage detector Cascade Mask RCNN \cite{cai2018cascade}, one-stage detector RetinaNet, and anchor-free detector RepPoints, with stronger backbone ResNeXt101 (X101) \cite{xie2017aggregated}.

The results in Tab. \ref{tab:strdet} demonstrate that, even if compared with stronger vanilla teachers, our method still can further improve the performance of distillation. For example, AKD brings \textbf{0.7 AP} gain on retinaNet for CWD and \textbf{0.5 AP} gain on RepPoints for MGD, which are practical and huge improvements to detection distillation. Besides, we find that AKD affords more support to detecting large size objects (AP$_{L}$), as larger objects would involve more data disturbance when generating Avatars. From another perspective, distillation of one-stage detectors is more sensitive to Avatar mechanism, where it brings about 0.4 to 0.7 AP gain, and we assume that it is uncertainty-aware weight that filters more semantic information, filling the gap with the two-stage detectors.

\definecolor{goldenrod}{RGB}{240,250,220}
\begin{table}[ht]
    \centering
    % \hspace{2mm}
    \begin{minipage}[t]{0.50\textwidth}
    \renewcommand\arraystretch{1.33}
	\setlength\tabcolsep{0.35mm}
	\centering
	\caption{\textbf{Object detection performance with stronger Avatars on COCO val set.} \textit{CM RCNN}: Cascade Mask RCNN.}
     %\vspace{-2mm}
	\label{tab:strdet}
	%\footnotesize
	\begin{tabular}{l|ccc|ccc}
	    \shline
	    Method & AP & AP$_{50}$ & AP$_{75}$ & AP$_{S}$ & AP$_{M}$ & AP$_{L}$\\
	    \shline
	    \scriptsize{T: CM RCNN-X101} & 45.6 & 64.1 & 49.7 & 26.2 & 49.6 & 60.0 \\
	    \scriptsize{S: Faster RCNN-R50} & 38.4 & 59.0 & 42.0 & 21.5 & 42.1 & 50.3 \\
        CWD~\cite{shu2021channel} & 41.7 & 62.0 & 45.5 & 23.3 & 45.5 & 55.5\\
	    \rowcolor{goldenrod}
	    + AKD & \textbf{42.0} &  \textbf{62.3} & \textbf{45.7} & \textbf{23.6}& \textbf{45.9} & \textbf{55.7}\\
        MGD~\cite{yang2022masked} & 42.1 & 62.0 & 45.9 & 23.7 & 46.4 & 56.1\\
	    \rowcolor{goldenrod}
	    + AKD & \textbf{42.4} & \textbf{62.4} & \textbf{46.1} & \textbf{23.7}& \textbf{46.8} & \textbf{56.5}\\
	    \shline

	    T: RetinaNet-X101 & 41.2 & 62.1 & 45.1 & 24.0 & 45.5 & 53.5\\
	    S: RetinaNet-R50 & 37.4 & 56.7 & 39.6 & 20.0 & 40.7 & 49.7 \\
        CWD~\cite{shu2021channel} & 40.8 & 60.4 & 43.4 & 22.7 & 44.5 & 55.3\\
	    \rowcolor{goldenrod}
	    + AKD & \textbf{41.5} & \textbf{60.8} & \textbf{44.4} & \textbf{22.9}& \textbf{45.9} & \textbf{55.5}\\
        MGD~\cite{yang2022masked} & 41.0 & 60.3 & 43.8 & 23.4 & 45.3 & 55.7\\
	    \rowcolor{goldenrod}
	    + AKD & \textbf{41.4} & \textbf{60.7} & \textbf{44.1} & \textbf{23.5}& \textbf{45.9} & \textbf{55.9}\\
	    \shline

        T: RepPoints-X101 & 44.2 & 65.5 & 47.8 & 26.2 & 48.4 & 58.5\\
	    S: RepPoints-R50 & 38.6 & 59.6 & 41.6 & 22.5 & 42.2 & 50.4\\
        CWD~\cite{shu2021channel} & 42.0 & 63.0 & 45.3 & 24.1 & 46.1 & 55.0\\
	    \rowcolor{goldenrod}
	    + AKD & \textbf{42.3} & \textbf{63.1} & \textbf{45.4} & \textbf{24.1}& \textbf{46.4} & \textbf{55.9}\\
        MGD~\cite{yang2022masked} & 42.3 & 62.8 & 45.6 & 24.4 & 46.2 & 55.9\\
	    \rowcolor{goldenrod}
	    + AKD & \textbf{42.8} & \textbf{63.5} & \textbf{46.1} & \textbf{24.5}& \textbf{47.0} & \textbf{57.3}\\
	    \shline
     
	\end{tabular}
    \vspace{-2mm}
    \end{minipage}
\end{table}

\subsection{Semantic segmentation}
\subsubsection{\textbf{Datasets.}}
 We conduct experiments on Cityscapes dataset \cite{Cordts2016Cityscapes} to validate Avatars' teaching ability, which contains 5000 high-quality images (2975, 500, and 1525 images for the training, validation, and testing). We evaluate all the student networks with mean Intersection-over-Union (mIoU).

\subsubsection{\textbf{Network architectures.}}
For all segmentation experiments, we take PSPNet-R101 \cite{zhao2017pyramid} as the teacher network. While for the students, we use various frameworks (DeepLabV3 \cite{chen2018encoder} and PSPNet) with ResNet-18 (R18) to valid the effectiveness of our method.

\label{sec:ss-im}

\subsubsection{\textbf{Implementation Details.}}
For the semantic segmentation task, the feature maps of Avatars source from last feature map in the backbone of teacher. We conduct experiments following CWD and MasKD. All the models are trained with the official strategies of 40K iterations schedule with 512 × 1024 input size in MMSegmentation \cite{mmseg2020}, where the optimizer is SGD and the weight decay is 5e-4. A polynomial annealing learning rate scheduler is adopted with an initial value of 0.02. We run all the models on 8 V100 GPUs.

\subsubsection{\textbf{Experimental results}}
As shown in Tab. \ref{tab:seg}, AKD further improves the performance of state-of-the-art distillation methods for semantic segmentation. Both the homogeneous and heterogeneous Avatar distillation bring the students significant improvements, \textit{e.g.}, the ResNet-18 based PSPNet gets at most \textbf{6.15 mIoU} improvement and that based DeepLabV3 gets \textbf{3.22 mIoU}. Notably, although the increase of DeepLabV3-R18 is lower than that of PSPNet, AKD still can bring at least 0.42 mIoU gain even if MGD already distills for DeepLabV3 perfectly.

\definecolor{goldenrod}{RGB}{240,250,220}
\begin{table}[t]
    \centering
    % \hspace{2mm}
    \begin{minipage}[t]{0.50\textwidth}
    \renewcommand\arraystretch{1.33}
	\setlength\tabcolsep{0.35mm}
	\centering
	\caption{\textbf{Semantic segmentation performance with Avatars on Cityscapes val set.} FLOPs is measured based on an input size of 512 × 1024.}
     %\vspace{-2mm}
	\label{tab:seg}
	%\footnotesize
    \begin{tabular}{l|c|c|c}
        \shline
        {Method} & Params(M) & FLOPs(G) & {mIoU (\%)} \\
        \shline
        T: PSPNet-R101 & 70.43 & 574.9 & 78.34\\
        \hline
        \scriptsize{S: PSPNet-R18} & \multirow{5}*{12.9} & \multirow{5}*{507.4} & 69.85 \\
        CWD \cite{shu2021channel} & ~ & ~ & 72.74 \\
        \rowcolor{goldenrod}
	    + AKD & 12.9 & 507.4 &  \textbf{74.57} \\
        MGD \cite{yang2022masked} & ~ & ~ & 74.46 \\
        \rowcolor{goldenrod}
	    + AKD &  ~ & ~ & \textbf{76.00} \\
        \hline
        \scriptsize{S: DeepLabV3-R18} & \multirow{5}*{13.6} & \multirow{5}*{572.0} & 73.20 \\
        CWD \cite{shu2021channel} & ~ & ~ & 74.34 \\
        \rowcolor{goldenrod}
	    + AKD & 13.6 & 572.0 & \textbf{75.02} \\
        MGD \cite{yang2022masked} & ~ & ~ & 76.02 \\
        \rowcolor{goldenrod}
	    + AKD &  ~ & ~ & \textbf{76.42} \\
        \shline
	\end{tabular}
     \vspace{-2mm}
    \end{minipage}
\end{table}

%------------------------------------------------------------------------
\section{Analysis}

% table (Ablation of components in MSD)
\begin{table}[ht]
    \centering
    \begin{minipage}[c]{0.50\textwidth}
    \renewcommand\arraystretch{1.4}
	\setlength\tabcolsep{0.35mm}
	\centering
	%\vspace{-4mm}
	\caption{\textbf{Ablation of components in AKD.} We train the student \textit{RetinaNet-R50} with teacher \textit{RetinaNet-X101} using 2$\times$ schedule, and adopts CWD (based on KL) and MGD (based on MSE) distillation on feature.}
     %\vspace{-2mm}
	\label{tab:ab_component}
	%\footnotesize
	\begin{tabular}{c|cc|c}
	    \shline
	    Method & Avatars & Uncertainty & AP(\%) \\
	    \shline
        ~ & \xmark & \xmark & 40.8 (baseline)\\
	    CWD (KL) & \cmark & \xmark & 41.0 (+0.2 AP)\\
	    ~ & \cellcolor{goldenrod}\cmark & \cellcolor{goldenrod}\cmark & \cellcolor{goldenrod}41.5 (+0.7 AP)\\
	    \shline
        ~ & \xmark & \xmark & 41.0 (baseline)\\
	    MGD (MSE)& \cmark & \xmark & 41.2 (+0.2 AP)\\
	    ~ & \cellcolor{goldenrod}\cmark & \cellcolor{goldenrod}\cmark & \cellcolor{goldenrod}41.4 (+0.4 AP)\\
	    \shline
	\end{tabular}
     %\vspace{-2mm}
	\end{minipage}\hfill
\end{table}

\subsection{Why Uncertainty for Evaluation Quality}

\begin{table}[h]

    \renewcommand\arraystretch{1.45}
	\setlength\tabcolsep{0.65mm}
	\centering
	%\begin{center}
	\caption{\textbf{The comparison of unequal training module on COCO (AP) with CWD.} Teacher: RetinaNet-X101. Student: RetinaNet-R50.}
    \vspace{-2mm}	
 \label{tab:unequal training module}
	%\footnotesize
	%\scriptsize
	%\normalsize
	\begin{tabular}{cc|ccca}
	    \shline
	    Teacher & Student & \thead{SE module\\\cite{hu2018squeeze}} & \thead{SAM module\\\cite{zhu2019empirical}} & \thead{CBAM module\\ \cite{woo2018cbam}} & \thead{Uncertainty\\ \textbf{Ours}}  \\
	    \shline
	    41.2 & 37.4 & 41.1 & 40.9 & 41.1 & \textbf{41.5} \\
	    \shline
	\end{tabular}
     %\vspace{-2mm}
	%\end{center}
\end{table}

To validate our uncertainty-aware weight effectiveness, we choose several common unequal training modules (SE \cite{hu2018squeeze}, SAM \cite{zhu2019empirical}, and CBAM \cite{woo2018cbam}) to watch Avatars distillation. The results are reported on Tab. \ref{tab:unequal training module}. We find that discriminating Avatars from channel or spatial or both of them dimension indeed works (0.1 AP), but the improvement they brought is far lower than uncertainty-aware weight brought. Our uncertainty-aware weight models the noise from Avatars generation, and force “mutant” Avatars to lower their effects to the student, instead of simple information redistribution. What's more, as shown in Tab. \ref{tab:ab_component}, the uncertainty-aware weight improves our Avatar mechanism by +0.2 AP, it well restrains the negative Avatars during the random generation.

\subsection{How to Merge Uncertainty}
\label{sec:merge}
\begin{table}[h]
    \centering
    % \begin{minipage}[c]{0.47\textwidth}
    \renewcommand\arraystretch{1.4}
	\setlength\tabcolsep{0.35mm}
	\centering
	% \vspace{-4mm}
	\caption{\textbf{Results of uncertainty mergence in different dimensions.} We train the student \textit{RetinaNet-R50} with teacher \textit{RetinaNet-X101} using 2$\times$ schedule and adopts CWD as distillation term. Chan.: Channel-wise, Spat.: Spatial-wise and Bat.: Batch-wise.}
	\label{tab:merge}
	%\footnotesize
	\begin{tabular}{cc|c|c|c}
	    \shline
	    Chan. & Spat. & $ \sigma^2 $ Across & $ \sigma^2 $ shape & AP(\%) \\
	    \shline
        \cmark & \cmark & Bat. + Chan. + Spat. & $ 1 \times 1 \times 1 $ & 40.8 (baseline) \\
        \xmark & \xmark & Bat. & $ C \times H \times W $ & 41.0 (+0.2 AP) \\
	    \cellcolor{goldenrod} \xmark & \cellcolor{goldenrod} \cmark & \cellcolor{goldenrod} Bat. + Spat. & \cellcolor{goldenrod} $ C \times 1 \times 1 $ & \cellcolor{goldenrod} 41.5 (+0.7 AP) \\
	    \cmark & \xmark & Bat. + Chan. & $ 1 \times H \times W $ & 41.2 (+0.4 AP) \\
        
	    \shline
	\end{tabular}
    % \vspace{-4mm}
	% \end{minipage}\hfill
\end{table}

In our previous statement in Sec.~\ref{sec:uncertainty}, mergence of uncertainty in certain dimensions provides a more stable form for distillation. We now discuss how to merge uncertainty simply from channel-wise and spatial-wise perspectives with Tab.~\ref{tab:merge}. Merging uncertainty in all dimensions gives an fixed scalar, which is equivalent to fixed temperature in CWD and serves as baseline. Comparing with baseline, using uncertainty across batch and spatial yields the best performance with 41.5 AP. And other two solutions (uncertainty across batch and across batch + channel + spatial) report 0.2 and 0.4 AP improvement over baseline. However, these two solutions all suffer from $ \sigma^2 $ shape mismatching problems when networks are trained with multi-scale images. To address it, interpolation has to be leveraged. Consequently it will lose some information in $ \sigma^2 $ tensor and hurt the performance. But the consistent improvements over baseline still reveal the effectiveness of the proposed uncertainty-aware weight.

\subsection{Visualization}

    \begin{figure}
        \centering
        \includegraphics[width=0.49\textwidth]{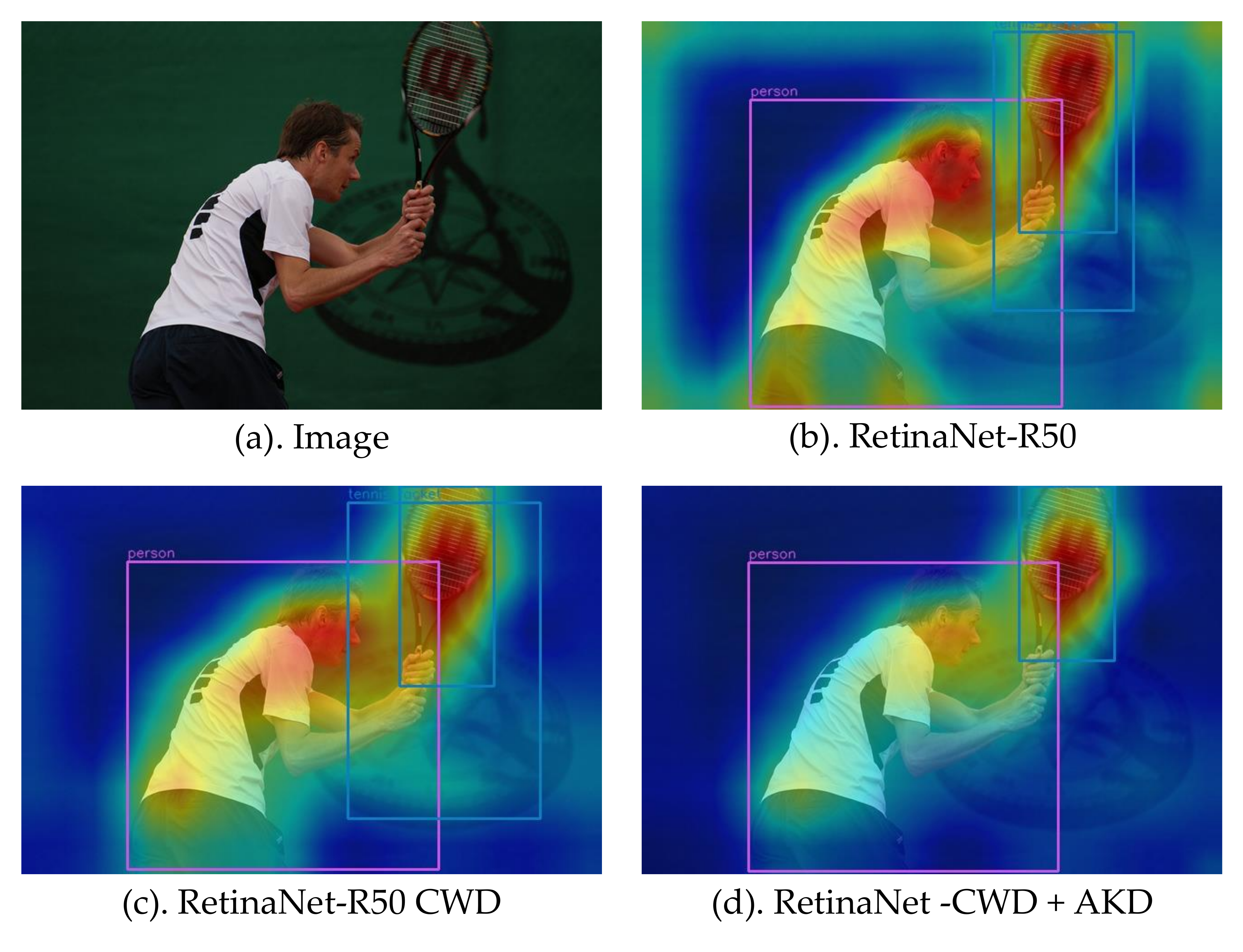}
        %\vspace{-6mm}
        \caption{\textbf{Comparison of feature maps with baselines.} We compare our method with RetinaNet-R50 (w/o distillation), RetinaNet-R50 distilled by RetinaNet-X101 with CWD. The image is randomly selected from val set of COCO dataset and the heatmaps are generated with AblationCAM.}
        \label{fig:cam}
        \vspace{-2mm}
    \end{figure}

    We visualize the heatmaps generated by AblationCAM~\cite{ramaswamy2020ablation} in Fig.~\ref{fig:cam} to further investigate effect of AKD. In the image, a player is swinging his tennis racket. RetinaNet-R50 detects the player and his tennis racket successfully. However, the model confusingly spares attention to four corners and generates redundant bounding box of racket. After distillation, RetinaNet-R50 CWD tackles attention problem. It focuses more on the foreground, but still detects two racket by mistake. On this basis, our method yields no repeat bounding box. When comparing heat maps of RetinaNet-R50 CWD and RetinaNet-CWD + AKD, the latter spare less attention on the player and seems to perform worse than the former. However, in fact, our method further gathers attention on the player and racket rather than the background, especially the misleading logo on the ground. And as the result, AKD yields more clear contrast between background and foreground in heat map and it is important to produce the accurate bounding boxes in latter process.
    
% \subsection{Sensitivity study of hyper-parameters}

% table (loss weight | dropout ratio)

\section{Conclusions}

In this paper, we propose our novel \textbf{\textit{AvatarKD}} framework to achieve impressive distillation performance when only one teacher model is available. In AKD, we put forward a concept called \textbf{\textit{Avatars}}. Each Avatar stems from the same teacher and provides unique description of teacher output. Combining all Avatars of one teacher, various perspectives of knowledge are accessible, like multiple teachers. Moreover, \textbf{\textit{uncertainty-aware}} weight is introduced. It takes statistical differences between the teacher and Avatars into consideration and balances importance Avatars in distillation adaptively. To investigate the effect of AKD, extensive experiments are conducted on classification, object detection and semantic segmentation. And encouraging SOTA results are observed on COCO and Cityscapes datasets when compared with other distillation algorithms, validating our motivation and approach.

\textbf{Acknowledgement.} This work was supported by the grants from the National Natural Science Foundation of
China under contract No.U20A20204. Besides, we would like to acknowledge the technical support provided by Jianfei Gao in Shanghai AI Lab.
%------------------------------------------------------------------------
\newpage
\bibliographystyle{ACM-Reference-Format}
\balance
\bibliography{main}
\end{document}